\documentclass[letterpaper]{article} % DO NOT CHANGE THIS
\usepackage{aaai2026}  % DO NOT CHANGE THIS
\usepackage{times}  % DO NOT CHANGE THIS
\usepackage{helvet}  % DO NOT CHANGE THIS
\usepackage{courier}  % DO NOT CHANGE THIS
\usepackage[hyphens]{url}  % DO NOT CHANGE THIS
\usepackage{graphicx} % DO NOT CHANGE THIS
\usepackage{natbib}  % DO NOT CHANGE THIS AND DO NOT ADD ANY OPTIONS TO IT
\usepackage{caption} % DO NOT CHANGE THIS AND DO NOT ADD ANY OPTIONS TO IT
\usepackage{algorithm}
\usepackage{algorithmic}

\usepackage{multirow}
\usepackage{adjustbox}
\usepackage{booktabs}
\usepackage{amssymb}
\usepackage{xspace}
\usepackage{amsmath}
\usepackage{bm}
\usepackage{pifont}
\usepackage{overpic}

\usepackage{newfloat}
\usepackage{listings}
\DeclareCaptionStyle{ruled}{labelfont=normalfont,labelsep=colon,strut=off} % DO NOT CHANGE THIS
\floatstyle{ruled}
\newfloat{listing}{tb}{lst}{}
\floatname{listing}{Listing}
\newcounter{corfn}
\def\correspondingauthors{%
  \ifnum\value{corfn}=0%
    \footnote{Corresponding authors.}%
    \setcounter{corfn}{\value{footnote}}%
  \else%
    \footnotemark[\value{corfn}]%
  \fi%
}%

\newcommand{\ie}{\textit{i}.\textit{e}.}

\title{History-Enhanced Two-Stage Transformer for \\Aerial Vision-and-Language Navigation}
\author{
    Xichen Ding\textsuperscript{\rm 1,\rm 2}\equalcontrib
    Jianzhe Gao\textsuperscript{\rm 3}\equalcontrib,
    Cong Pan\textsuperscript{\rm 1,\rm 2},
    Wenguan Wang\textsuperscript{\rm 3}\correspondingauthors,
    Jie Qin\textsuperscript{\rm 1,\rm 2}\correspondingauthors
}
\affiliations{
    \textsuperscript{\rm 1}College of Artificial Intelligence, Nanjing University of Aeronautics
and Astronautics\\
    \textsuperscript{\rm 2}Key Laboratory of Brain-Machine Intelligence
Technology, Ministry of Education, China\\
    \textsuperscript{\rm 3}The State Key Lab of Brain-Machine Intelligence, Zhejiang University\\
}

\usepackage{bibentry}
\begin{document}

\maketitle

\begin{abstract}
Aerial Vision-and-Language Navigation (AVLN) requires Unmanned Aerial Vehicle (UAV) agents to localize targets in large-scale urban environments based on linguistic instructions. While successful navigation demands both global environmental reasoning and local scene comprehension, existing UAV agents typically adopt mono-granularity frameworks that struggle to balance these two aspects. To address this limitation, this work proposes a History-Enhanced Two-Stage Transformer (HETT) framework, which integrates the two aspects through a coarse-to-fine navigation pipeline. Specifically, HETT first predicts coarse-grained target positions by fusing spatial landmarks and historical context, then refines actions via fine-grained visual analysis. In addition, a historical grid map is designed to dynamically aggregate visual features into a structured spatial memory, enhancing comprehensive scene awareness. Additionally, the CityNav dataset annotations are manually refined to enhance data quality. Experiments on the refined CityNav dataset show that HETT delivers significant performance gains, while extensive ablation studies further verify the effectiveness of each component.

\end{abstract}

\begin{links}
    \link{Code \& Dataset}{https://github.com/crotonyl/HETT}
\end{links}

\section{Introduction}

\begin{figure}[!htb]
    \centering
    \includegraphics[width=1\linewidth]{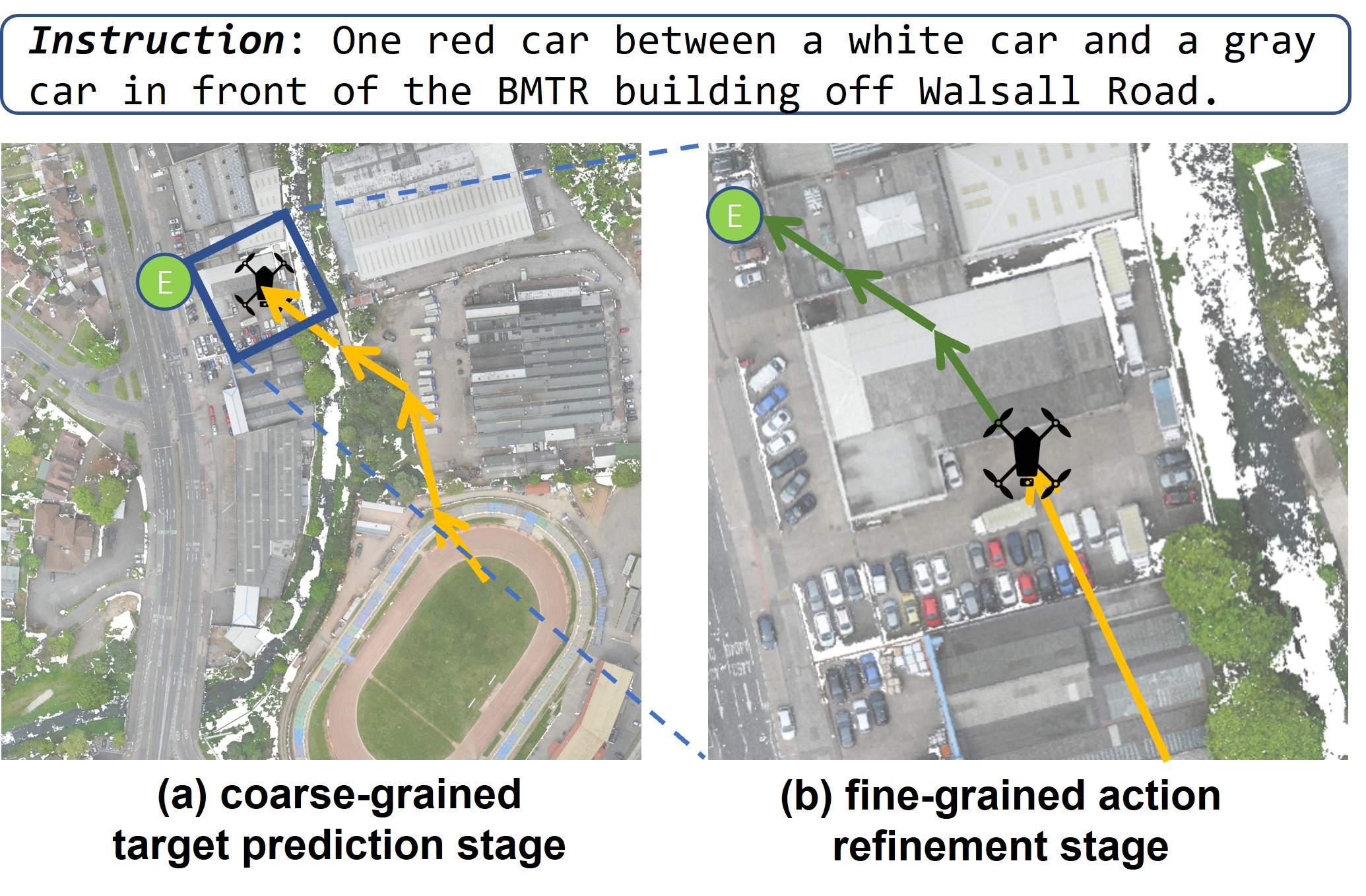}
    \caption{\textbf{Illustration} of HETT. Given a goal-oriented instruction, HETT operates in two stages: (a) our agent first leverages coarse-grained global features to predict a global target, providing high-level directional guidance for long-range navigation; (b) it then refines local actions on fine-grained local features to adapt to local observations.} 
    \label{fig:Policy}
\end{figure}

Aerial Vision-and-Language Navigation (AVLN) is an emerging challenge in embodied AI, requiring Unmanned Aerial Vehicle (UAV) agents to identify and locate targets in outdoor environments given natural-language instructions~\cite{fan-etal-2023-aerial,liu_2023_AerialVLN,lee2024citynavlanguagegoalaerialnavigation}. Unlike indoor Vision-and-Language Navigation (VLN) tasks~\cite{R2R,qi2020reverie,qiao2022hop}, which are confined to limited action spaces, AVLN requires UAV agents to navigate in unstructured and large-scale aerial environments. This poses critical challenges in sustaining robust cross-modal alignment between vision and language throughout long and dynamic trajectories.

To achieve robust AVLN performance, UAV agents need to integrate adaptive decision-making with continuous environmental understanding~\cite{10203831,chen2021hamt,qiao2022hop, w2}. However, current state-of-the-art AVLN agents \ding{182} \textit{primarily employ mono-granularity frameworks, and the integration of global planning and local perception remains underexplored}. Specifically, local path planning approaches~\cite{w8,w7,Su_An_Chen_Yu_Ning_Ling_Huang_Wang_2025, w1} focus on fine-grained alignment between local visual observations and instruction semantics, predicting actions within a predefined action space. Global path planning approaches~\cite{lee2024citynavlanguagegoalaerialnavigation,w3,wang2024realisticuavvisionlanguagenavigation}, in contrast, construct coarse-grained 2D spatial maps for target position prediction. While both paradigms contribute important capabilities to AVLN, they also present complementary limitations: local planning excels at dynamic adaptation but struggles with long-horizon reasoning due to its dependence on local perceptions, whereas global planning offers comprehensive spatial awareness but lacks the fine-grained visual understanding needed for precise localization. Furthermore, \ding{183} \textit{existing UAV agents fail to preserve historical details during long-term navigation planning}. Recent agents~\cite{gao2024stmraerialvln,lee2024citynavlanguagegoalaerialnavigation,w4} typically project semantic masks onto a top-down map using UAV pose and depth information to represent historical context. However, their global environmental comprehension remains heavily dependent on semantic segmentation modules~\cite{10204268} such as GroundingDINO~\cite{liu2023grounding} and SAM~\cite{kirillov2023segany}. This reliance fundamentally limits the capture of fine-grained visual details, potentially compromising overall scene understanding~\cite{wang2023gridmm,w5,w6}. Moreover, \ding{184} \textit{a key limitation of existing AVLN datasets lies in suboptimal annotation quality}. As the field is still in an early stage, many datasets~\cite{lee2024citynavlanguagegoalaerialnavigation} rely on LLM-generated navigation annotations without rigorous manual review, introducing noise and inaccuracies into the training data.

To address these challenges, this work proposes a History-Enhanced Two-Stage Transformer (HETT) for AVLN, which integrates coarse-grained and fine-grained multimodal information to bridge the gap between global planning and local perception. Motivated by \ding{182}, HETT adopts a two-stage navigation policy that decomposes the navigation process into \textit{coarse-grained target prediction} and \textit{fine-grained action refinement}. As illustrated in Fig.~\ref{fig:Policy}, during the first stage (Fig.~\ref{fig:Policy}(a)), our agent leverages prior spatial landmarks and accumulated historical information to infer the target’s approximate location. Once approaching this predicted region, the agent enters the second stage (Fig.~\ref{fig:Policy}(b)), where detailed visual cues guide precise local movements. For \ding{183}, our agent incorporates a historical grid map that encodes both temporal and spatial information of the globally visited environment. The map partitions the environment into uniformly sized grid cells, each storing fine-grained visual features based on their coordinates. This design enables persistent and structured historical memory across long navigation trajectories. In addition, regarding \ding{184}, we perform a thorough manual refinement of the dataset annotations to mitigate the noise introduced by LLM-generated annotations and ensure the reliability of our evaluation.

Experiments on the CityNav benchmark validate the effectiveness of HETT, showing substantial improvements of \textbf{14.16}\%, \textbf{10.75}\%, and \textbf{18.00}\% in SR across the validation and test sets. Ablation studies verify the contributions of our core components as well as the impact of dataset refinement.

\section{Related Work}
\noindent\textbf{Aerial Vision-and-Language Navigation (AVLN).} The widespread adoption of UAVs drives extensive research in AVLN, where drones navigate outdoor environments based on language instructions and visual observations. Seminal works include AVDN, which provides manually collected dialog-based instructions for AVLN tasks~\cite{fan-etal-2023-aerial}, and CityNav, which enhances navigation by incorporating GPS-augmented target descriptions~\cite{lee2024citynavlanguagegoalaerialnavigation}. Recent simulators further accelerate progress by constructing photorealistic 3D outdoor environments with full 6-DoF UAV control~\cite{liu_2023_AerialVLN,wang2024realisticuavvisionlanguagenavigation}.

Despite these developments, current AVLN agents still face several language-grounding challenges. Urban environments typically contain dense landmark distributions, irregular street layouts, and highly variable urban geometry, making it difficult for agents to maintain stable cross-modal alignment over long navigation trajectories. In addition, most existing agents lack effective mechanisms for modeling historical visual–linguistic context~\cite{gao2024stmraerialvln}, limiting their ability to resolve ambiguous or deferred references in multi-step navigation instructions. These limitations demonstrate the need for more advanced AVLN architectures that jointly integrate linguistic understanding with robust spatial reasoning in large-scale outdoor settings.

\noindent\textbf{Vision-and-Language Navigation (VLN).} VLN is a fundamental task in embodied AI in which agents navigate photorealistic scenes using natural-language instructions~\cite{gao20253dgs}. Representative benchmarks include R2R~\cite{8578485}, RxR~\cite{rxr}, and CVDN~\cite{thomason:arxiv19}, all constructed within indoor household environments in the Matterport3D simulator~\cite{Matterport3D}. These datasets primarily focus on discrete action spaces and set the foundation for indoor VLN evaluation. The introduction of VLN-CE~\cite{krantz2020navgraph} marks a major shift toward realism by converting topological trajectories into continuous action spaces, thereby reflecting real-world motion dynamics.

Traditional VLN agents rely on cross-modal attention mechanisms to align visual observations with textual commands. However, these agents often struggle to capture temporal dependencies, as they predominantly attend to the current observation while overlooking accumulated historical context~\cite{w10}. HAMT~\cite{chen2021hamt} introduces a hierarchical transformer that encodes the complete navigation history as sequential memory tokens. TD-STP~\cite{zhao2022target} extends HAMT by incorporating a target prediction mechanism that enables agents to ``imagine'' future states. DUET~\cite{Chen_2022_DUET} equips agents with topological map encoding to facilitate efficient global planning. Other agents~\cite{georgakis2022cm2, huang23vlmaps,w9} maintain a top-down semantic map to better capture the spatial layout and structural relations.

\noindent\textbf{Aerial Navigation.} AVLN builds upon the broader field of autonomous aerial navigation, which is traditionally organized around two complementary paradigms: global planning, where agents leverage environmental context for semantic goal inference, and local planning, where agents rely on immediate perceptual cues for reactive control.

Global navigation agents typically compute offline optimal routes from satellite imagery or Digital Elevation Models (DEMs). Early agents~\cite{869506} employ sparse A* search with spatial constraints to reduce computational overhead during long-distance route planning, while HGARL~\cite{10882533} demonstrates that hybrid metaheuristic agents based on HHO effectively avoid local minima in obstacle-dense environments. However, global agents remain limited by their dependence on static scene maps. In contrast, local agents prioritize real-time reactivity using high-frequency perceptual observations. Hrabar’s stereo-vision agent~\cite{4650775} achieves sub-meter obstacle avoidance in cluttered spaces through probabilistic roadmaps. Follow-up agents introduce artificial potential fields for dynamic obstacle avoidance and genetic-evolutionary strategies for optimizing 3D trajectories, significantly improving the agent's real-time adaptability.

\begin{figure}[t]
    \centering
    \includegraphics[width=0.95\linewidth]{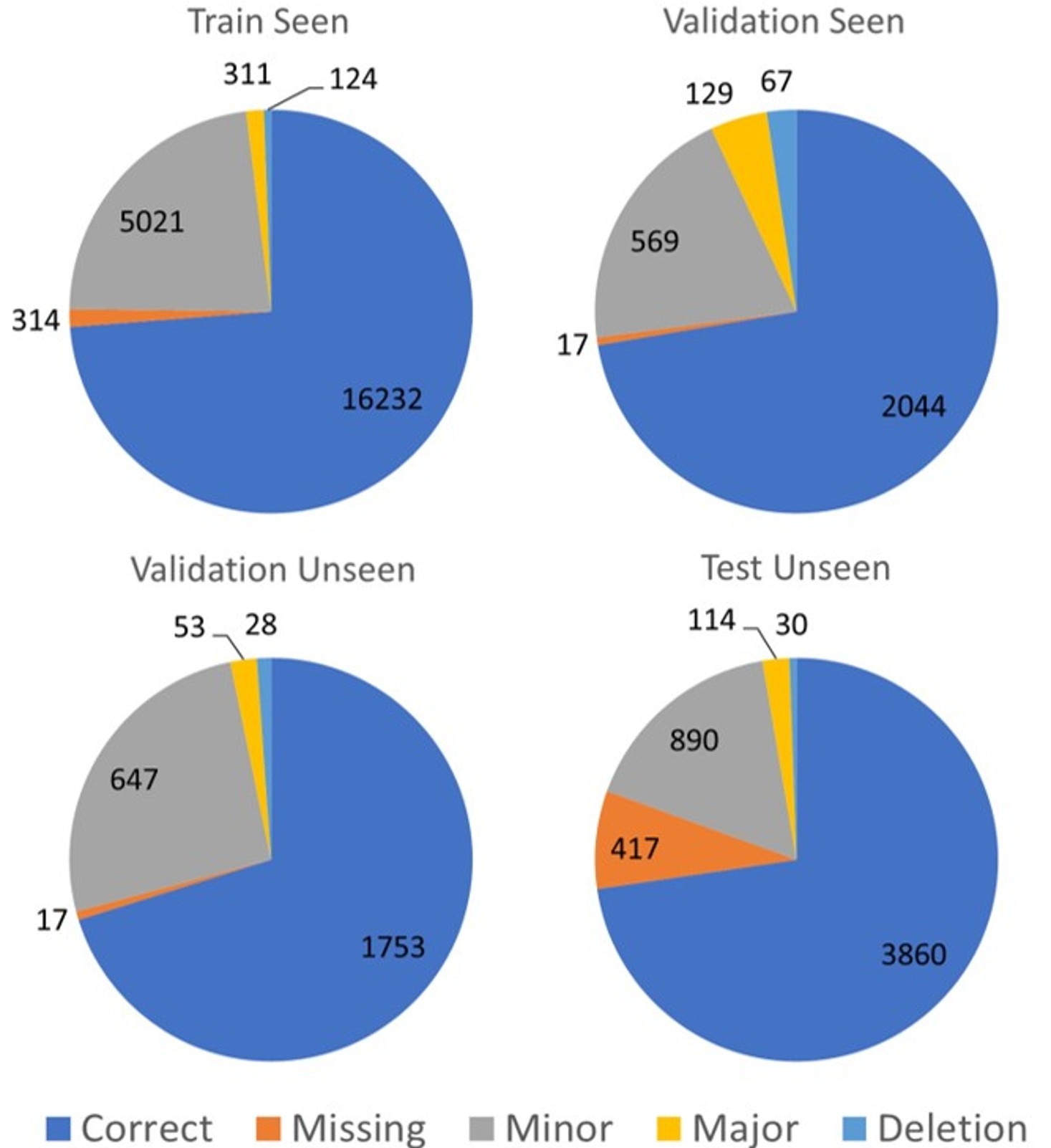}
\caption{\textbf{Annotation Error Distribution} in the CityNav~\cite{lee2024citynavlanguagegoalaerialnavigation} Dataset. The chart reports the proportion of each error type in the original annotations, including ``Missing'' landmark references, ``Minor'' textual inconsistencies, ``Major'' landmark extraction failures, and ``Deletion'' cases that lack usable spatial grounding.}

    \label{fig:statistics}
\end{figure}

\section{Dataset}

\label{sec:dataset}

\begin{table}[t]
\setlength{\tabcolsep}{2pt}
\centering
\begin{tabular}{r|c|c|c|c}
\hline
Types & Train Seen & Val Seen & Val Unseen & Test Unseen\\ 
\hline
\hline
Missing & 314 & 17 & 17 & 417\\
Minor & 5021 & 647 & 569 & 890\\
Major & 311 & 53 & 67 & 114\\
Deletion & 124 & 28 & 129 & 30\\

\hline
\end{tabular}
\caption{\textbf{Annotation Error Statistics} in the CityNav~\cite{lee2024citynavlanguagegoalaerialnavigation} Dataset. ``Missing'' indicates missing landmark references. ``Minor'' refers to spelling mistakes or other typos. ``Major'' denotes critical landmark extraction errors that misalign instructions with their intended targets. ``Deletion'' corresponds to instructions removed from the dataset due to lacking valid landmark references.}
\label{tab:dataset_refinement}
\end{table}

In AVLN, navigation instructions are commonly categorized into two paradigms: step-by-step instructions (\textit{e.g., ``Take off, fly through the tower of cable bridge and down to the end of the road.''}) and goal-oriented instructions (\textit{e.g., ``The white car that is the sixth car in the fifth aisle of the One Stop parking lot.''})~\cite{10203831}. The CityNav dataset~\cite{lee2024citynavlanguagegoalaerialnavigation} primarily utilizes goal-oriented instructions. Crucially, these instructions rely on pre-defined landmark information for target localization. Navigating unstructured urban environments under goal-oriented instructions necessitates a landmark-centric agent design, as precise localization is infeasible when relying solely on geometric cues or relative directions. Consequently, landmarks become essential spatial anchors that enable agents to position themselves and interpret high-level goal descriptions.

The original CityNav dataset contains 32K trajectories corresponding to natural language descriptions of approximately 5.8K objects such as buildings and cars. Its landmark annotations are initially generated using GPT-3.5 Turbo~\cite{ouyang2022traininglanguagemodelsfollow}. However, this automated process introduced substantial errors due to the absence of human verification. For instance, the model failed to correctly associate the landmark ``One Stop'' with the description: ``The white car that is the sixth car in the fifth aisle of the One Stop parking lot.'' To address such inaccuracies, we performed a manual refinement of the annotations, with quantitative statistics presented in Fig.~\ref{fig:statistics} and Table.~\ref{tab:dataset_refinement}. This ensures landmark correspondence for every instruction, thereby providing a reliable foundation for evaluation.

\section{Method}

\begin{figure*}[!tbh]
    \centering
    \includegraphics[width=1\textwidth]{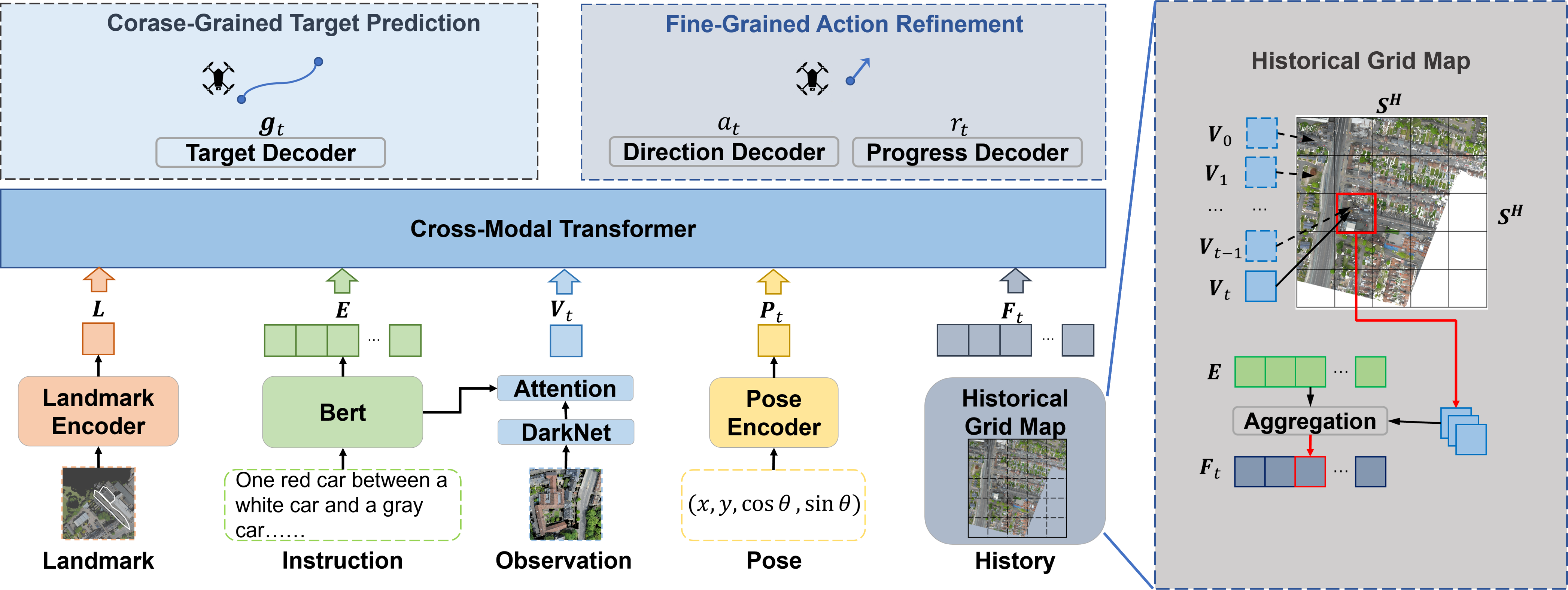}
    \caption{\textbf{Overview} of HETT. At time step $t$, five types of tokens (\ie, the landmark, instruction, pose, history, and view tokens) are sent into the cross-modal transformer to predict action decisions. In the \textit{Coarse-Grained Target Prediction} stage, our agent leverages the target prediction result~$\bm{g}_t$ to guide navigation. In the \textit{Fine-Grained Action Refinement} stage, the agent uses the local action estimation~$a_t$ to adjust immediate movements until the predicted progress~$r_t$ reaches threshold.}
    \label{fig:Framework}
\end{figure*}

\noindent\textbf{Problem Formulation.} In the CityNav benchmark~\cite{lee2024citynavlanguagegoalaerialnavigation}, a UAV agent navigates a 2D urban environment under the guidance of a natural-language instruction. At each time step, the agent receives an egocentric top-down RGB-D observation together with its current pose. The agent also has access to static geographic priors that provide polygonal boundary descriptions of landmarks mentioned in the instruction. Formally, given the instruction, the associated landmark priors, and the sequence of observations, the agent must generate a sequence of continuous control actions that drives it toward the instructed goal. A navigation episode is considered successful if the agent issues the [stop] action within 20 steps and the final predicted location falls within 20\,m of the ground-truth target.

\noindent\textbf{Overview (Fig.~\ref{fig:Framework}).} HETT integrates multi-modal cues to produce an adaptive navigation policy with coherent long-horizon reasoning. Given an instruction, text embeddings $\bm{E}$ are extracted, and the referenced landmark contours are encoded as spatial features $\bm{L}\!\in\!\mathbb{R}^{D}$. At each time step $t$, the agent obtains its observation and pose, which are encoded into visual features $\bm{V}_t\!\in\!\mathbb{R}^{D}$ and pose features $\bm{P}_t\!\in\!\mathbb{R}^{D}$. Historical environmental information is preserved through a Historical Grid Map that aggregates past visual features into a spatial memory tensor $\bm{F}_t\!\in\!\mathbb{R}^{D}$. These components are jointly processed by a transformer to yield fused contextual representations combining linguistic guidance, landmark priors, visual perception, pose state, and accumulated memory. Based on these, HETT operates in two stages. The coarse-grained stage predicts a target location $\bm{g}_t\!\in\!\mathbb[0,1]^2$ that provides high-level directional cues, while the fine-grained stage refines immediate movements through an action estimate $a_t\!\in\!(-\pi,\pi]$ and a progress indicator $r_t\!\in\![0,1]$, enabling precise and adaptive navigation.

\subsection{Two-Stage Transformer Framework}
To bridge the gap between long-horizon reasoning and fine-grained scene comprehension in AVLN, a Two-Stage Transformer Framework is introduced for AVLN. In the \textit{Coarse-Grained Target Prediction} stage, the UAV agent infers an approximate target location by leveraging pre-defined landmark information together with a coarse historical grid map. In the subsequent \textit{Fine-Grained Action Refinement} stage, the agent attends to local visual features to produce precise continuous actions for accurate trajectory execution.

\noindent\textbf{Coarse-Grained Target Prediction.} To rapidly narrow down the potential target region, the agent first builds a coarse spatial representation of the environment. Given the landmark priors referenced in the instruction, their polygonal contours are projected onto a top-down landmark map $\mathcal{M}^L\!\in\!\mathbb{R}^{S^L\times S^L}$, where $S^L$ is the map resolution. The projected contours are encoded into a landmark embedding $\bm{L}$:
\begin{equation}
\small
\begin{aligned}
\bm{L}\!=\!{\texttt{MLP}}\big(\texttt{CNN}(\mathcal{M}^L)\big)\!\in\!\mathbb{R}^{D}.
\end{aligned}
\label{eq_landmark}
\end{equation}

At time step $t$, the agent extracts visual features $\bm{V}_{t}\!\in\!\mathbb{R}^{D}$ from the current RGB-D observation using a pretrained Darknet-53 encoder~\cite{redmon2018yolov3incrementalimprovement}. To retain spatial and temporal context, a \textit{historical grid map} aggregates the sequence of visual features into a structured spatial memory $\bm{F}_t\!\in\!\mathbb{R}^{D}$. A \textit{Coarse-Grained Target Prediction} module then estimates the normalized global target position by jointly reasoning over the instruction embedding $\bm{E}\!\in\!\mathbb{R}^{N\times d}$, where $N$ is the instruction length, the landmark embedding $\bm{L}$, and the historical memory $\bm{F}_t$. These are jointly fused through a multi-layer transformer (\texttt{MLT}):
\begin{equation}\small
\begin{aligned}
\bm{G}_t \;=\; \texttt{MLT}\big([\bm{E};\bm{L};\bm{F}_t]\big)\;\in\;\mathbb{R}^{D},
\end{aligned}
\label{eq_FGT}
\end{equation}
where $[\,;\,]$ denotes feature concatenation. The fused representation $\bm{G}_t$ captures the essential spatial and visual cues required for global target inference, and the normalized target coordinates $\bm{g}_t$ are obtained as:
\begin{equation}\small
\bm{g}_t=\texttt{Softmax}\big(\texttt{MLP}(\bm{G}_t)\big)\in[0,1]^2.
\label{eq_targetprediction}
\end{equation}

\noindent\textbf{Fine-Grained Action Refinement.} Although the coarse-grained stage offers high-level directional guidance by estimating the target region, precise navigation further requires fine-grained alignment between linguistic instructions and visual observations. To achieve this, the agent enters the \textit{Fine-Grained Action Refinement} stage, which emphasizes detailed scene interpretation and accurate motion control. A cross-modal attention mechanism is applied to derive instruction-aware visual embeddings:
\begin{equation}
\small
\begin{aligned}
\bm{V}_t=\texttt{Attention}\big([\bm{E};\bm{O}_t]\big)\in\!\mathbb{R}^{D}.
\end{aligned}
\label{eq_visualencoding}
\end{equation}
where $\bm{O}_t$ denotes the visual feature map extracted from the top-down RGB-D observation, and $\bm{V}_t$ represents the refined visual embedding aligned with the instruction semantics. $\bm{V}_t$ are subsequently fused with the landmark embedding $\bm{L}$, historical spatial memory $\bm{F}_t$, and pose embedding $\bm{P}_t$:
\begin{equation}
\small
\bm{R}_t,\,\bm{A}_t = 
\texttt{MLT}\big([\bm{E};\,\bm{L};\,\bm{F}_t;\,\bm{V}_t;\,\bm{P}_t]\big)
\in\mathbb{R}^{D},
\label{eq_FVT}
\end{equation}
where $\bm{R}_t$ encodes the contextualized representation for action reasoning, and $\bm{A}_t$ serves as the basis for fine-grained action refinement in the subsequent control module. Based on these, fine-grained navigation actions are generated as:
\begin{equation}
\small
r_t=\texttt{Sigmoid}(\texttt{MLP}(\bm{R}_t))\in[0,1],
\label{eq_progress}
\end{equation}
\begin{equation}
\small
a_t=\texttt{Arctan2}\!\big(\texttt{Tanh}(\texttt{MLP}(\bm{A}_t))\big)\in(-\pi,\pi],
\label{eq_localaction}
\end{equation}
where $r_t$ provides an estimate of task completion progress for deciding when to terminate the episode, and $a_t$ denotes the turning angle used for immediate motion adjustment. The agent repeatedly executes this process until the predicted progress $r_t$ surpasses a predefined threshold or the maximum step limit is reached, enabling a balance between precise goal attainment and efficient trajectory completion.

\begin{table*}[!tbh]
\centering
\begin{tabular}{r|cccc|cccc|cccc}
\hline

\multirow{2}{*}{Models} & \multicolumn{4}{c|}{Validation Seen} & \multicolumn{4}{c|}{Validation Unseen} & \multicolumn{4}{c}{Test Unseen} \\ 
~ & NE$\downarrow$ & SR$\uparrow$ & OSR$\uparrow$ & SPL$\uparrow$ 
       & NE$\downarrow$ & SR$\uparrow$ & OSR$\uparrow$ & SPL$\uparrow$ 
       & NE$\downarrow$ & SR$\uparrow$ & OSR$\uparrow$ & SPL$\uparrow$ \\ 
\hline
\hline
Random          & 222.3 & 0.00  &  1.15 & 0.00  & 223.0 & 0.00 & 0.90  & 0.00 & 208.8 & 0.00 & 1.44  & 0.00 \\
Human & 9.1 & 89.31 & 96.40 & 60.17 & 9.4 & 88.39 & 95.54 & 62.66 & 9.8 & 87.86 & 95.29 & 57.04 \\
Seq2Seq    & 257.1 & 1.81  &  7.89 & 1.58  & 317.4 & 0.79 & 8.82  & 0.61 & 245.3 & 1.50 & 8.34  & 1.30 \\
CMA        & 240.8 & 0.95  &  9.42 & 0.92  & 268.8 & 0.65 & 7.86  & 0.63 & 252.6 & 0.82 & 9.70  & 0.79 \\
AerialVLN & 65.6  & 9.77  & 23.77 & 8.64  & 81.8  & 6.79 & 17.91 & 5.73 & 64.1 & 8.09 & 19.13 & 5.91\\
MGP & 53.0  & 16.93  & 29.90 & 14.38  & 73.8  & 8.35 & 17.91 & 7.07 & 86.1 & 10.90 & 20.24 & 9.94 \\

\hline
HETT(Ours) & 45.2 & 25.16 & 48.40 & 23.01 & 62.1 & 17.48 & 25.09 & 14.46 & 72.9 & 22.97 & 39.30 & 17.01\\
\textbf{HETT(Ours)*} & \textbf{37.2} & \textbf{31.09} & \textbf{51.86} & \textbf{25.76} & \textbf{51.3} & \textbf{19.10} & \textbf{34.78} & \textbf{16.70} & \textbf{40.4} & \textbf{28.90} & \textbf{49.56} & \textbf{23.79}\\
\hline
\end{tabular}
\caption{
{\textbf{Quantitative results} of HETT. HETT(Ours)* is trained and evaluated on the refined dataset.}
}
\label{tab:overall_results}
\end{table*}

\subsection{Historical Grid Map}
Inspired by history-encoding mechanisms in indoor VLN agents~\cite{chen2021hamt,wang2023gridmm,liu2024volumetricenvironmentrepresentationvisionlanguage}, a \textit{Historical Grid Map} is introduced to capture and organize the agent’s accumulated visual memories. The environment is discretized into a fixed $S^H\times S^H$ grid covering the entire navigation region. At each time step $t$, the agent stores its fine-grained visual features and their corresponding spatial coordinates into the historical map $\mathcal{M}^H_{t}$:
\begin{align}
\mathcal{M}^H_{t}= \mathcal{M}^H_{t-1} \cup [\bm{V}_t,\bm{p}_{t}],
\end{align}
where $\bm{p}_{t}$ represents the agent’s position.

These stored features are then assigned to their corresponding grid cells according to spatial coordinates, forming a structured grid feature set:
\begin{equation}
\begin{aligned}
\mathcal{M}^H_{t,(x,y)} = \{\bm{m}_{t,j}\}_{j=1}^{J}, 
\quad (x,y)\in\{1,\ldots,S^H\}^2,
\end{aligned}
\label{eq_gtprogress}
\end{equation}
where each $\bm{m}_{t,j}\!\in\! \mathbb{R}^{D}$ denotes a visual feature whose spatial coordinate falls inside the grid cell indexed by $(x,y)$, and $J$ is the number of features accumulated in that cell. 

For each cell, a relevance matrix $\bm{K}_{(x,y)}$ is computed between its feature set $\mathcal{M}^H_{t,(x,y)}$ and instruction embedding $\bm{E}$:
\begin{equation}
\bm{K}_{(x,y)}
    = \texttt{Softmax}\!\big(\mathcal{M}^H_{t,(x,y)} \cdot\bm{E}^\top\big)
    \in [0,1]^{J\times N}.
\end{equation}
where $N$ denotes the length of the instruction.
Finally, the historical grid token at cell $(x,y)$ is computed via a relevance-weighted aggregation:
\begin{equation}
\bm{F}_{t,(x,y)}
    = \sum\nolimits_{j=1}^{J} K_{(x,y),\,j} \cdot \bm{m}_{t,j}
    \in \mathbb{R}^{D},
\label{eq_gridtoken}
\end{equation}
where $K_{(x,y),\,j}$ denotes the scalar relevance weight associated with feature $\bm{m}_{t,j}$. Aggregating across all grid cells yields the full structured spatial memory 
$\bm{F}_t\!\in\!\mathbb{R}^{D}$.

\begin{figure*}[t]
    \centering
    \includegraphics[width=1\linewidth]{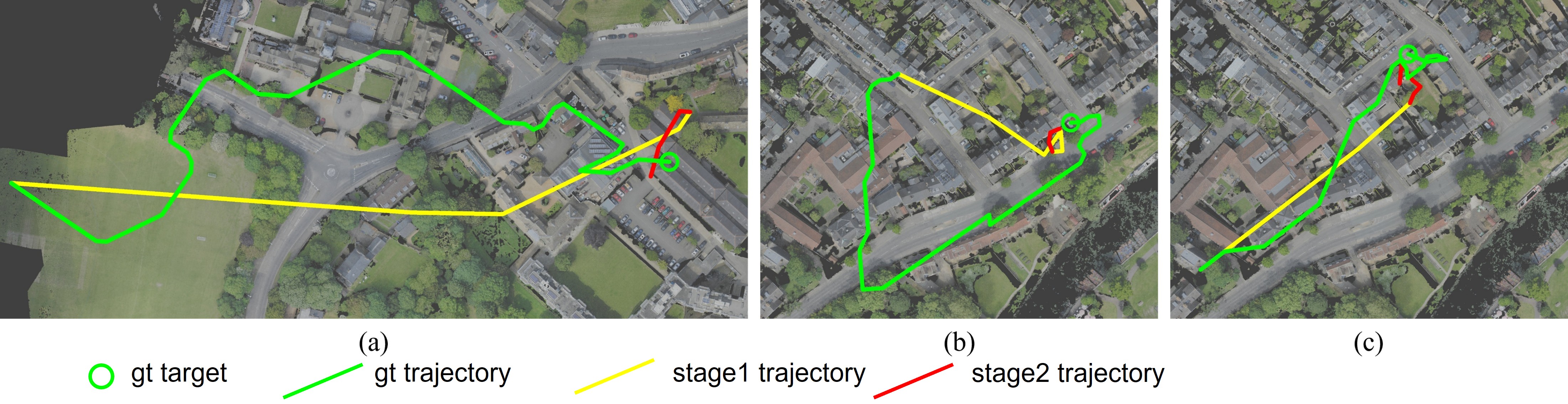}
    \caption{\textbf{Visualization results} for the two-stage navigation. Initial predictions in the coarse-grained stage may exhibit target drift or premature stopping. The fine-grained refinement stage corrects these deviations and steers the agent toward the ground-truth target. Three representative cases are visualized to illustrate this behavior.}
    \label{fig:stage_vis}
\end{figure*}

\subsection{Loss Function}
Following prior works~\cite{Chen_2022_DUET,Hong_2022_CVPR,zhao2022target}, DAgger~\cite{ross2011reduction} is adopted for policy training. To supervise the proposed two-stage framework, three dedicated loss functions are introduced. The first component is the coarse-grained target prediction loss:
\begin{equation}
\begin{aligned}
\mathcal{L}^G\!&=\sum\nolimits_{t=1}^{T}\texttt{MSE}(\bm{g}_t,\bm{g}^{gt}),
\end{aligned}
\label{lossg}
\end{equation}
where $\bm{g}_t$ is the predicted normalized target coordinate and $\bm{g}^{gt}$ denotes the corresponding ground-truth target location. Similarly, the action loss $\mathcal{L}^A$ and progress loss $\mathcal{L}^R$ are formulated using the ground-truth action $a^{gt}_t$ and ground-truth progress $r^{gt}_t$, respectively.

The total training objective is expressed as:
\begin{equation}
\small
\mathcal{L}
= \alpha_{1}\mathcal{L}^{G}
+ \alpha_{2}\mathcal{L}^{A}
+ \alpha_{3}\mathcal{L}^{R},
\label{eq_losssum}
\end{equation}
where $\alpha_{\{1,2,3\}}$ are weighting coefficients that balance the contributions of the three losses.

\subsection{Implementation Details}

Our model is implemented in PyTorch and trained on four 24GB RTX A5000 GPUs for 20 epochs, with a batch size of 2, a learning rate of 1e-4, and AdamW optimizer. The grid size $d$ is set to 5. $\alpha_1, \alpha_2, \alpha_3$ are set to 2.0, 1.5, 0.1.

\section{Experiment}
\subsection{Experimental Setup}
\noindent\textbf{Dataset Preparation.}
The navigation instructions utilized in our experiments are derived from the original and refined CityNav~\cite{lee2024citynavlanguagegoalaerialnavigation} dataset, which comprises 32,326 natural language descriptions paired with human demonstration trajectories, all collected by crowd-sourcing. Each language description is rich in detail, encompassing information such as landmarks, regions and objects, etc. The drone images are taken from SensatUrban, which gathers orthographic projections and depth maps of 13 blocks in Birmingham and 33 blocks in Cambridge. These data are utilized to simulate the RGBD input an actual drone would acquire during navigation. Additionally, the geometric outlines of landmarks within the geographic information database are obtained from CityRefer~\cite{miyanishi_2023_NeurIPS}, providing essential information for the navigation tasks.

\noindent\textbf{Evaluation Metrics.}
To comprehensively evaluate the navigation performance of our HETT, we adopt four standard metrics commonly used in the field: Navigation Error (NE), Success Rate (SR), Oracle Success Rate (OSR), and Success weighted by Path Length (SPL).
\begin{itemize}
    \item Navigation Error (NE): This metric quantifies the Euclidean distance between the final position of the UAV agent and the ground truth target location.
    \item Success Rate (SR): This metric calculates the percentage of episodes where the UAV agent terminates its navigation within a pre-defined success threshold.
    \item Oracle Success Rate (OSR): This metric assesses whether the agent's trajectory at any point approaches the target within the success threshold.
    \item Success weighted by Path Length (SPL): This metric is the success rate weighted by the ratio of the reference path length to the actual path length traveled by the agent.
\end{itemize}

\begin{table*}[h]

\center
\begin{tabular}{c|ccc|ccc|ccc}
\hline
\multirow{2}{*}{\#} &   \multirow{2}{*}{Dataset} &\multirow{2}{*}{Two-Stage}   & \multirow{2}{*}{History} & \multicolumn{3}{c|}{Validation Seen} & \multicolumn{3}{c}{Validation Unseen}  \\
~ & & & & NE$\downarrow$ & SR$\uparrow$ & OSR$\uparrow$ 
       & NE$\downarrow$ & SR$\uparrow$ & OSR$\uparrow$  \\ 
\hline
\hline
1 & - & - & - &   49.66 & 19.42 & 33.23 & 65.84 & 9.84 & 20.24  \\ 
  
2 & \checkmark & - & - &
39.72 & 24.98 &  42.59   &  56.32  & 13.27   & 29.11 \\

3 &   \checkmark  & \checkmark & -
& 40.47 & 26.19 &   48.58   &  56.33  &  14.68 & 34.71   \\ 

4 & \checkmark & - & \checkmark  
& \textbf{36.38} & 29.31  &   41.78   &  52.99  & 15.28  & 26.88    \\

\textbf{5} & \checkmark & \checkmark & \checkmark  
& 37.24 & \textbf{31.09} & \textbf{51.86}   &  \textbf{51.34} & \textbf{19.10} & \textbf{34.78}    \\
\hline

\end{tabular}
\caption{\textbf{Ablated results} of the main components on the CityNav dataset.}

\label{tab:main-ablation}
\end{table*}

\begin{table}
\setlength{\tabcolsep}{3pt}
\center
\begin{tabular}{c|ccc|ccc}
\hline
\multirow{2}{*}{Agent} &  \multicolumn{3}{c|}{Validation Seen} & \multicolumn{3}{c}{Validation Unseen}  \\
~ & NE$\downarrow$ & SR$\uparrow$ & OSR$\uparrow$ 
       & NE$\downarrow$ & SR$\uparrow$ & OSR$\uparrow$  \\ 
\hline\hline
AerialVLN & 65.6  & 9.77  & \textbf{23.77}  & 81.8  & 6.79 & 17.41\\
\textbf{AerialVLN*} & \textbf{54.2}  & \textbf{12.38}  & 22.22  & \textbf{65.9}  & \textbf{9.12} & \textbf{17.72}\\
\hline
MGP & 53.0  & 16.93  & 29.90 & 73.8  & 8.35 & 17.91 \\
\textbf{MGP*} & \textbf{48.1}  & \textbf{19.17}  & \textbf{35.51} & \textbf{66.5}  & \textbf{10.47} & \textbf{28.06} \\

\hline
\end{tabular}
\caption{\textbf{Ablated results} of dataset refinement. * denotes models trained and evaluated on the refined CityNav dataset.}

\label{tab:dataset-ablation}
\end{table}

\subsection{Overall Performance}\label{exp_overall}
\noindent\textbf{Quantitative Results.} We compare HETT with several baseline UAV agents on the CityNav benchmark. As shown in Table~\ref{tab:overall_results}, HETT achieves consistent improvements across all metrics and data splits. In terms of SR, it surpasses the strongest baseline MGP by \textbf{8.23}\%, \textbf{9.13}\%, and \textbf{12.07}\% on the validation-seen, validation-unseen, and test-unseen sets, respectively. Beyond SR, HETT also shows notable gains on the test-unseen split, reducing NE by 13.2 m, improving OSR by 19.06\%, and increasing SPL by 7.07\%. To further examine the benefit of our refined annotations, we train a variant denoted as HETT*, where * indicates that both training and evaluation are performed on the refined dataset. HETT* obtains additional SR improvements of \textbf{5.93}\%, \textbf{1.62}\%, and \textbf{5.93}\% across the three splits, along with corresponding reductions in NE and increases in SPL. These consistent improvements clearly indicate that the proposed HETT framework yields more accurate and stable UAV navigation, while the refined dataset annotations further enhance the reliability of supervision signals for AVLN.

\begin{figure}[t]
    \centering
    \includegraphics[width=1\linewidth]{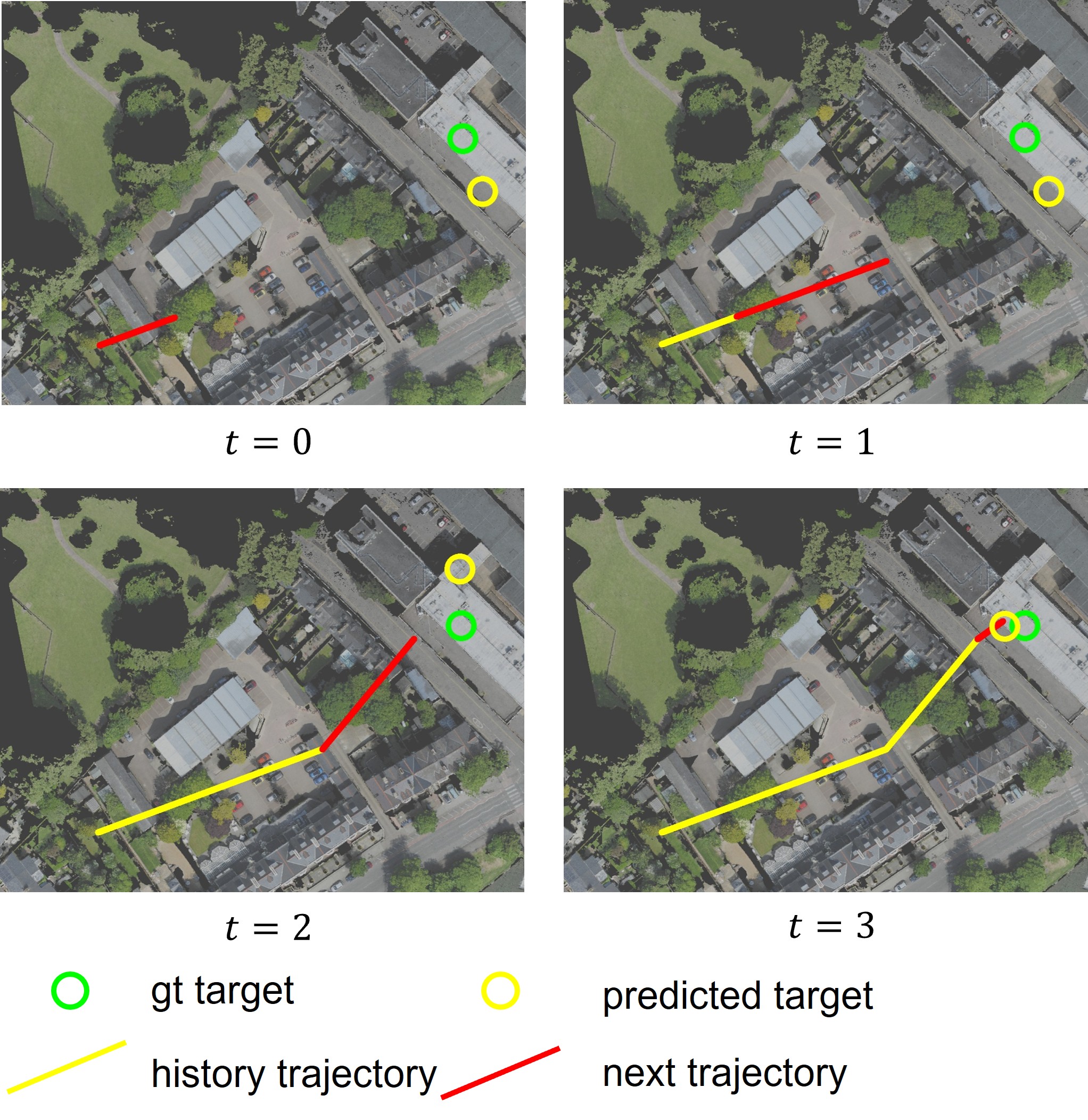}
    \caption{\textbf{Visualization} of the Coarse-Grained Target Prediction stage. As navigation proceeds, the predicted position steadily converges toward the ground-truth location.}
    \label{fig:stage1_vis}
\end{figure}

\noindent\textbf{Qualitative Results.} Fig.~\ref{fig:stage_vis} illustrates how HETT executes the \textbf{two-stage navigation policy}. In the \textbf{coarse-grained target prediction} stage, the agent first moves toward the estimated target region based on global spatial cues. Once it reaches this vicinity, the agent switches to the \textbf{fine-grained action refinement} stage, where localized perception and historical context are leveraged to perform precise trajectory adjustments. As shown in cases (a) and (b), the refined actions correct deviations accumulated during the coarse prediction stage, while in case (c), they guide the agent to achieve accurate final alignment with the ground-truth target. These qualitative examples clearly demonstrate the adaptiveness and effectiveness of our strategy.

To further assess the impact of the \textbf{Historical Grid Map}, Fig.~\ref{fig:stage1_vis} visualizes the evolution of target predictions across navigation steps. The predicted target position transitions from an initially uncertain estimate to a stable convergence around the ground-truth location as more observations are accumulated. This progression highlights that this map successfully incorporates temporal visual cues, thereby enhancing global localization accuracy.
\subsection{Ablation Studies}
To validate the contributions of key components of our HETT, we perform systematic ablation studies. The results presented in Tables \ref{tab:main-ablation}, \ref{tab:dataset-ablation}, and \ref{tab:grid-ablation} quantitatively assess the effectiveness of individual modules.

\noindent\textbf{Effectiveness of Each Component.}
We begin by evaluating the contribution of each core component in our framework. As shown in Table~\ref{tab:main-ablation}, incorporating the \textbf{refined dataset} (Row~\#2) leads to a clear performance improvement over the baseline in Row~\#1, increasing SR from 19.42\% to \textbf{24.98}\% on the validation-seen split and from 9.84\% to \textbf{13.27}\% on the validation-unseen split. The effect of the \textbf{two-stage navigation policy} is examined by comparing Row~\#3 with Row~\#2. With SR improving from 24.98\% to \textbf{26.19}\% on the validation-seen split and from 13.27\% to \textbf{14.68}\% on the validation-unseen split, the two-stage navigation policy further enhances both long-horizon reasoning and local action accuracy. Moreover, the \textbf{historical grid map} provides the most substantial gains. Compared with Row~\#2, adding the historical grid map (Row~\#4) improves SR from 24.98\% to \textbf{29.31}\% on the validation-seen split and from 13.27\% to \textbf{15.28}\% on the validation-unseen split. When combined with the two-stage policy (Row~\#5 \textit{vs.} Row~\#3), SR increases from 26.19\% to \textbf{31.09}\% on validation seen and from 14.68\% to \textbf{19.10}\% on validation unseen. These show that each component contributes meaningfully to overall performance.

\noindent\textbf{Analysis of Dataset Refinement.} We further evaluate the influence of dataset refinement on other UAV agents. As shown in Table~\ref{tab:dataset-ablation}, both MGP and AerialVLN trained on the refined annotations achieve consistently higher performance than their original counterparts. On the Validation Seen split, AerialVLN improves from 9.77\% to \textbf{12.38}\%, while MGP rises from 16.93\% to \textbf{19.17}\%. Similar gains are observed on the Validation Unseen split, where AerialVLN increases from 6.79\% to \textbf{9.12}\%, and MGP from 8.35\% to \textbf{10.47}\%. These results further validate the effectiveness of the refined dataset and highlight the importance of high-quality landmark annotations for robust AVLN training and evaluation.

\begin{table}[t]
\setlength{\tabcolsep}{5pt}
\center
\begin{tabular}{c|ccc|ccc}
\hline
\multirow{2}{*}{$S^H\!\times\!S^H$} &  \multicolumn{3}{c|}{Validation Seen} & \multicolumn{3}{c}{Validation Unseen}  \\
~ & NE$\downarrow$ & SR$\uparrow$ & OSR$\uparrow$ 
       & NE$\downarrow$ & SR$\uparrow$ & OSR$\uparrow$  \\ 
\hline\hline

{0$\times$0} & 40.47 & 26.19 &   48.58   &  56.33  &  14.68 & 34.71 \\

{3$\times$3} & 39.92 & 27.87 & \textbf{52.11} & 53.98 & 18.61 & \textbf{36.41}   \\ 

{\textbf{5$\times$5}} & \textbf{37.24} & \textbf{31.09}  &   51.86   &  \textbf{51.34}  & \textbf{19.10}  & 34.78 \\

{7$\times$7} & 37.28 & 27.45  &   41.26   &  54.43  & 17.87  & 32.81    \\

\hline
\end{tabular}
\caption{\textbf{Ablated results} of the grid size ($S^H\!\times\!S^H$) in Historical Grid Map.}

\label{tab:grid-ablation}
\end{table}

\noindent\textbf{Analysis of Historical Grid Map.} Furthermore, we ablate the historical grid map’s grid size. The results are summarized in Table \ref{tab:grid-ablation}, where a $0\times0$ grid indicates the absence of the historical grid map. The experimental results show that the model with a $3\times3$ grid size achieves the optimal OSR of \textbf{52.11}\%, \textbf{36.41}\% on the validation seen and unseen set. When the grid size increases to $5\times5$, the model attains the best overall performance with SR rising to \textbf{31.09}\% and \textbf{19.10}\% respectively on the validation seen and unseen set; however, when the grid size reaches $7\times7$, both SR and OSR on the validation unseen set decrease, likely due to excessive features interfering with the model’s ability to extract discriminative navigation cues. Thus, we select the $5\times5$ configuration as the optimal grid size for our final model.

\section{Conclusion}

In this paper, we propose a History-Enhanced Two-Stage Transformer (HETT) for AVLN. HETT adopts a coarse-to-fine navigation paradigm that decomposes the navigation process into a two-stage navigation policy: coarse-grained target prediction and fine-grained action refinement. Moreover, the historical grid map further enhances the agent's spatial awareness by maintaining structured environmental memory during navigation. Compared with previous UAV agents, our HETT integrates both coarse-grained environmental perception and fine-grained visual cues, thus enabling more accurate navigation results. In addition, we conduct manual refinement of the CityNav annotations, providing a more reliable benchmark for AVLN. Extensive experiments demonstrate the effectiveness of our HETT. one limitation of HETT is its dependency on pre-defined information. Future work will investigate online environment mapping to enhance navigation robustness.

\bibliography{aaai2026}

\end{document}